# Exploiting problem structure in a genetic algorithm approach to a nurse rostering problem.

Uwe Aickelin
School of Computer Science
University of Nottingham
NG8 1BB   UK
uxa@cs.nott.ac.uk

Kathryn A. Dowsland
European Business Management School
University of Wales Swansea
Singleton Park
Swansea SA2 8PP
k.a.dowsland@swan.ac.uk

**Summary.**
There is considerable interest in the use of genetic algorithms to solve problems arising in the areas of scheduling and timetabling. However, the classical genetic algorithm paradigm is not well equipped to handle the conflict between objectives and constraints that typically occurs in such problems. In order to overcome this, successful implementations frequently make use of problem specific knowledge. This paper is concerned with the development of a GA for a nurse rostering problem at a major UK hospital. The structure of the constraints is used as the basis for a co-evolutionary strategy using co-operating sub-populations. Problem specific knowledge is also used to define a system of incentives and disincentives, and a complementary mutation operator. Empirical results based on 52 weeks of live data show how these features are able to improve an unsuccessful canonical GA to the point where it is able to provide a practical solution to the problem.

# Exploiting problem structure in a genetic algorithm approach to a nurse rostering problem.

## 1. Introduction

In recent years, genetic algorithms (GAs) have emerged as a useful tool for the heuristic solution of complex discrete optimisation problems. In particular, there has been considerable interest in their use in the solution of scheduling and timetabling problems, for example [3], [8], [20] and [23]. However, such problems frequently involve conflicting objectives and constraints, which tend to cause problems for the classical GA paradigm. Hence, successful implementations require mechanisms for overcoming these difficulties, and these can often be provided by exploiting problem specific knowledge. This paper examines these issues in the context of a GA approach to the solution of a nurse scheduling problem arising at a major UK hospital. A simple canonical GA is unable to deal with the constraints in the problem, even when violations are penalised adaptively. This is remedied by the incorporation of three types of problem specific knowledge. The most effective of these is a co-evolutionary strategy using co-operating sub-populations. The objective of the research was to investigate the efficacy of such knowledge, rather than to produce the best possible evolutionary algorithm approach to the problem. Nevertheless, the final implementation is able to provide high quality solutions in less than a minute of computation time on a standard Pentium II 200 PC.

The paper starts with a detailed description of the problem and the canonical GA that forms the basis of our algorithm. The shortcomings of this implementation are highlighted and a co-evolutionary strategy, based on a partition of the population into a number of co-operating sub-populations, is developed to overcome these. Empirical results are used to illustrate the efficacy of this approach. We then show how solutions can be further improved by a combination of rewards, penalties and intelligent mutations, designed to guide the search towards promising solutions and to help them fulfil their potential. We conclude with some observations as to the applicability of such modifications to other scheduling problems.

## 2. The Nurse Scheduling Problem and IP Model

Our problem is that of creating weekly schedules for wards of up to 30 nurses at a major UK hospital. These schedules have to satisfy working contracts and meet the demand for given numbers of nurses of different grades on each shift, while being seen to be fair by the staff concerned. The latter objective is achieved by meeting as many of the nurses' requests as possible and by considering historical information to ensure that unsatisfied requests and unpopular shifts are evenly distributed. The day is partitioned into three shifts: two day shifts known as 'earlies' and 'lates', and a longer night shift. Due to hospital policy, a nurse would normally work either days or nights in a given week, and because of the difference in shift length, a full week's work would normally include more days than nights. For example, a full time nurse would work 5 days or 4 nights, whereas typical part-time contracts are for 4 days or 3 nights, 3 days or 3 nights and 3 days or 2 nights.

As described in [7], the problem can be decomposed into three independent stages. The first uses a knapsack model to ensure that sufficient nurses are available to meet the covering constraints. If not, additional 'bank' nurses are allocated to the ward, so that the problem tackled in the second phase is always feasible. The second stage is the most difficult and involves allocating the actual days or nights to be worked by each nurse. Once this has been decided a third phase uses a network flow model to allocate those on days to 'earlies' and 'lates'. As stages 1 and 3 can be solved quickly using standard knapsack and network flow algorithms this paper is only concerned with the highly constrained second step.

This part of the problem can be modelled as follows. Each possible shift pattern worked by a given nurse can be represented as a zero-one vector with 14 elements, where the first seven elements represent the seven days of the week and the last seven the corresponding nights. A one in the vector denotes a scheduled day or night on and a zero a day or night off. Depending on the working hours of a nurse there are a limited number of shift patterns available to her / him. For instance, a full time nurse working either 5 days or 4 nights has a total of 21 (i.e. $\binom{7}{5}$) feasible day shift patterns and 35 (i.e. $\binom{7}{4}$) feasible night shift patterns. Typically, there will be between 20 and 30 nurses per ward, 3 grade-bands, 9 part time options and 411 different shift patterns. Depending on the nurses' preferences, the recent history of patterns worked, and the overall attractiveness of the pattern, a penalty cost is allocated to each nurse-shift pattern pair. These values were set in close consultation with the hospital and range from 0 (perfect) to 100 (unacceptable), with a bias to lower values. Further details can be found in [7].

The second stage problem can be formulated as an integer linear program as follows.

Decision variables:
$$x_{ij} = \begin{cases} 1 & \text{nurse i works shift pattern j} \\ 0 & \text{else} \end{cases}$$

Parameters:
n = number of nurses
m = number of possible shift patterns
p = number of grades
$$a_{jk} = \begin{cases} 1 & \text{pattern j covers day k} \\ 0 & \text{else} \end{cases}$$
$$q_{is} = \begin{cases} 1 & \text{nurse i is of grade s or higher} \\ 0 & \text{else} \end{cases}$$
$p_{ij}$ = penalty cost of nurse i working shift pattern j
$R_{ks}$ = demand of nurses of grade s or above on day/night k
F(i) = set of feasible shift patterns for nurse i

Target function:
$$\sum_{i=1}^{n} \sum_{j=1}^{m} p_{ij} x_{ij} \rightarrow \min! \quad (1)$$

s.t.

$$\sum_{j \in F(i)} x_{ij} = 1 \quad \forall i \quad (2)$$

$$\sum_{j=1}^{m} \sum_{i=1}^{n} q_{is} a_{jk} x_{ij} \geq R_{ks} \quad \forall k,s \quad (3)$$

Constraints (2) ensure that every nurse works exactly one shift pattern from her feasible set. Constraints (3) ensure that the demand for nurses is fulfilled for every grade on every day/night. Note that the definition of $q_{is}$ ensures that higher graded nurses can substitute those at lower grades if necessary. This problem can be regarded as a multiple-choice set-covering problem. The sets are given by the shift pattern vectors and the objective is to minimise the cost of the sets needed to provide sufficient cover for each shift at each grade. The multiple-choice aspect derives from constraints (2), which enforce the choice of exactly one pattern (or set) from the alternatives available for each nurse.

This problem is NP-hard [12], [13] and our instances typically involve between 1000 and 2000 variables and up to 70 constraints. The difficulty of a particular instance will depend on the shape of the solution landscape, which in turn depends on the distribution of the penalty weights and their relationship with the set of solutions meeting the covering requirements as given in (3). An examination of 52 real data sets based on three different wards at different times of the year showed that these characteristics differed widely between different problem instances. While there were some easy instances with many low cost global optima scattered liberally throughout the solution space, others had few global optima, and in some cases feasible solutions were also relatively sparse. The effects of this variety is borne out by experiments using other techniques. Using a standard IP package a few instances were solved at the initial node, but three remained unsolved after each being allowed 15 hours run-time on a Pentium II 200. Experiments with a number of descent methods using different neighbourhoods, and a standard simulated annealing implementation, were even less successful and frequently failed to find feasible solutions. The most successful approach to date is based on tabu search [7]. However, the quality of solutions relies heavily on chains of moves that work well because of the way in which the different factors affecting the quality of a schedule are combined into the $p_{ij}$ values. As this will differ from hospital to hospital, the goal of a fast but robust approach that is able to perform well on instances based on other derivations of the $p_{ij}$ values is worth pursuing. Here we develop a GA approach to the problem. The next section outlines the basic features of a GA, and the following section gives details of our basic implementation.

**3. Genetic Algorithms.**

Genetic algorithms are heuristics based on the principles of natural evolution and 'survival of the fittest'. Solutions are represented by an encoding as a string, analogous to chromosomes in genetics, and are usually referred to as individuals. The algorithm starts with a number of solutions, the so-called population, which is usually randomly generated. Then a series of genetic operators (selection, crossover, mutation and replacement) are applied to the solutions in the population to produce a new population. One full sequence of these operators is called a generation and the process is continued for a number of generations until a stopping criterion is met.

The 'survival of the fittest' principle guides the search towards good solutions as follows. A suitable target function is used to measure the fitness of each solution. The 'fitter' a solution, the more likely it is to be chosen in the selection stage to contribute to new solutions. In our algorithm, the selection is based on rank rather than on absolute fitness to avoid scaling problems (cf. [21], [22]). New solutions are formed by the crossover operation, where two of the chosen solutions are picked and recombined to form new solutions. A small proportion of these new solutions are then mutated, i.e. changed slightly in a random way. Once an appropriate number of new solutions have been created, they replace an equivalent number of old solutions, with some of the fitter old solutions surviving to the next generation.

The genetic algorithm scheme can be summarised as follows:

1. Encoding of solutions and initialisation of first population.
2. Fitness evaluation of all solutions in the population.
3. Selection of the parent solutions according to their fitness.
4. Crossover between parents to form new solutions.
5. Mutation of a small proportion of these new solutions.
6. Fitness evaluation of new solutions.
7. Replacement of old solutions by new ones keeping some of the (best) old solutions.
8. Returning to step 3 if stopping criteria are unfulfilled.

The following section describes how this basic paradigm was implemented for our nurse-scheduling problem.

**4. Coding and Constraints**

There are many different possibilities for encoding the problem. Here we use an encoding that follows directly from the IP formulation in Section 2 and is equivalent to that used by Hadj-Alouane and Bean [10] for general multiple-choice integer programs. Each individual represents a full one-week schedule, i.e. it is a string of *n* elements with *n* being the number of nurses. The *ith* element of the string is the index of the shift pattern worked by nurse *i*. For example, if we have 5 nurses, the string (1,17,56,67,3) represents the schedule in which nurse 1 works pattern 1, nurse 2 pattern 17 etc.. Such an encoding means that any standard crossover allocates some nurses to the shift patterns worked in one parent solution and the remainder to those worked in the other parent. Here we use a uniform crossover (i.e. a binary mask of length equal to that of the string is generated randomly, and positions with a 1 in the mask are copied from the one parent and those corresponding to a zero are copied from the other.) Our mutation operator changes the shift-pattern of just one nurse. As each nurse corresponds to one position in the string, this ensures that all child solutions obey the constraints that each nurse works exactly one shift pattern. As long as the initial values, and those allowed for mutation of each element, are selected from the values of feasible patterns for each particular nurse, then constraint set (2) is automatically taken care of by the encoding. This leaves only constraints (3) to be dealt with further.

In principle, there are three ways of dealing with constraints in a GA: implementation into the encoding / crossover, repair operators, and the use of penalty functions [14], [22]. All three have been used successfully in the solution of the classical set-covering problem (i.e. set covering without multiple-choice constraints). Wren and Wren [23] take the first approach and ensure feasibility of the children by using a special form of crossover that pools the sets used in the parents and produces one or more feasible children from these sets. Beasley and Chu [2] opt for the use of a repair operator that adds additional elements to any infeasible children until the covering constraints are satisfied. An example of the penalty approach is that of Richardson et al. [18] who reduce fitness by an amount proportional to the number of unsatisfied constraints. However, all but the penalty approach rely on the fact that feasible solutions can always be obtained from infeasible ones by adding additional sets or elements. Where this is not the case none of these methods is directly transferable. In our problem it is the multiple-choice constraints that prevent us simply adding extra shift patterns to an infeasible solution. This suggests that infeasible individuals will need to be present in the solution space. We therefore chose a penalty approach similar to that in [18] which reduces the fitness in proportion to the amount of violation for each constraint violated.

The disadvantages of this approach are the problem of finding a suitable penalty term and the possibility that the GA may terminate without finding a feasible solution. The former problem is addressed by Hadj-Alouane and Bean [10], who consider a GA approach to a multiple-choice integer program occurring in facility location model. They point out that in problems where there are many optimal solutions, the lower bound obtained by relaxing constraints (3) in Lagrangian fashion is frequently slack. This implies that even if a different penalty is associated with each violated constraint there is no combination of linear penalties that gives a relaxation equivalent to the original problem. In view of [10] we considered both linear and quadratic penalties for our problem. However, as there appeared to be no advantage in the use of quadratic penalties the remainder of the paper is based on the use of the linear form.

Using a linear penalty function with $g_{demand}$ as the penalty weight and the same notation as in section 2, the fitness function becomes:

$$\sum_{i=1}^{n}\sum_{j=1}^{m}p_{ij}x_{ij} + g_{demand}\sum_{k=1}^{14}\sum_{s=1}^{p}\max\left[\left(R_{ks} - \sum_{i=1}^{n}\sum_{j=1}^{m}q_{is}a_{jk}x_{ij}\right); 0\right]$$

This value is equal to the raw fitness of a solution or individual. However, as stated earlier we use rank based selection throughout our experiments. Thus the raw fitness was used to calculate the ranks.

For our initial implementations other parameters were based on those suggested in popular genetic algorithm literature [9], [22] as given below.

| | | |
|---|---|---|
| Population size | = | 1000 |
| Weight for uncovered demand | = | 10 |
| Selection of solutions for crossover | = | based on rank in population |
| Number of parents/children per crossover | = | 2 |
| Chance of a single bit uniform crossover= | 0.75 | |
| Mutation rate | = | 2% |
| Surviving solutions in each generation | = | best 10% of current population |
| Stopping criteria | = | 30 generations without improvement |

With the exception of the penalty weight, these were retained after a series of parameter tuning experiments. For the penalty weights we follow the advice of Reeves [17] who points out that several researchers have found that the use of adaptive or dynamic penalty weights can overcome the problems observed with fixed weights. For example Smith and Tate [19] suggest a weight that is scaled according to the difference between the fitness of the best feasible solution and the overall fittest solution found so far. Hadj-Alouane and Bean [10] start with a high penalty and reduce it when all solutions have been infeasible for several generations, increasing it again when the best solution over several generations has been feasible. Our adaptive weight is based on the number of violated constraints in the best solution to date and is given by the following expression.

$$g_{demand} = \begin{cases} \alpha \cdot q & \text{for } q > 0 \\ v & \text{for } q = 0 \end{cases}$$

where $q$ is the number of constraints violated by the best solution, $\alpha$ is a pre-set severity parameter, and $v$ is a suitably small value. Thus the weight depends on the number of violated constraints until a feasible solution is found, after which it remains at $v$. As a result of experimentation $\alpha$ and $v$ were set at 8 and 5 respectively.

## 5. Computational experiments.

The development work for our algorithm was based on the 52 real data sets obtained from 3 different wards and covering different times of the year. All these problems have been solved to completion via the IP formulation, and the optimal solution is therefore known. The performance of the tabu search algorithm, with and without the special chain moves, is also available for this data. Once we were satisfied with the performance of the GA on these data sets, our final implementation was also tested on a further 52 problems with the same requirements and nurses as the original but with the $p_{ij}$ values generated randomly. Once again the results were compared with those of the tabu search. In each of the GA experiments the algorithm was run 20 times using different starting populations and different random number streams.

The results using the basic algorithm with the parameters given above and the dynamic penalty weights are shown in Figure 1a. These results provide a basis by which to measure any improvements derived from the introduction of problem specific information. The bars above the y axis represent solution quality, with the black bars showing the number of optimal solutions, and the total bar height showing the number of solutions within 3 units of the optimal value. The value of 3 was chosen as it corresponds to the penalty cost of violating the least important level of requests in the original formulation. Thus solutions this close to the optimum would certainly be acceptable to the hospital. The bars below the axis represent the number of times out of 20 that the run terminated without finding a single feasible solution. Thus the less the area below the axis and the more above, the better the performance. Note that for 7 of the problems all 20 runs failed to find any feasible solutions. Just 21 problems resulted in at least one solution in the acceptable range and there were only two optimal terminations out of all 1040 runs. In the next section we suggest one possible reason for this poor performance and show how this can be overcome by the introduction of a series of co-operative sub-populations.

*Insert Fig 1 about here.*

## 6. Co-operating Sub-Populations.

The success of a genetic algorithm is usually attributed to the validity of the 'Building Block Hypothesis' [11]. This relies on the crossover operator being able to combine good partial solutions, so-called building blocks, into complete good solutions. However, it has long been recognised (see for example Davidor [4], Davis[5]) that for problems where there is a high degree of epistasis, or non-linearity, this is not guaranteed. Although the objective function as given by (1) is linear in the costs of the patterns chosen for each nurse, the inclusion of penalty function results in a highly epsistatic problem. The source of this non-linearity is two-fold. The main effect derives from the fact that the contribution to cover of a nurse working a particular shift depends on the patterns worked by the other nurses. But, because higher graded nurses are allowed to cover for those of lower grades, there is a second dimension to this interdependency. It is this second source that our co-evolutionary approach is intended to reduce.

Our approach was motivated by an observation made during parameter testing. We noted that one-point crossover sometimes gave better results than uniform, leading us to conjecture that in some cases combining large building blocks, made up of good partial solutions for one or more grades, aided the solution process. Beasley et al [1] suggest that this sort of effect can be achieved by decomposing the problem into several sub-problems and then overlaying the sub-solutions. Here we take a similar approach, but instead of overlaying sub-solutions we adopt the ideas underpinning a parallel GA [15], [21] and attempt to breed sub-populations that are highly fit with respect to nurses within specific grades. Our sub-populations are subjected to a strategy of co-operative co-evolution similar to that presented by Potter and De Jong [16]. They suggest using species representing sub-components of a potential solution, and evolving each species using a standard GA. Complete solutions are formed by amalgamating the components, and these completions are used to calculate fitness. Our approach differs in that the fitness of an individual in a particular species is based solely on the sub-components represented by that species. Complete solutions are achieved by maintaining a hierarchy of species so that species at level 2 are based on the sub-components of two of the species at level 1 etc. A proportion of each generation at higher levels is produced by strategic combination of parents from lower levels.

This process can be summarised as follows.

1. Sort the solution strings according to nurses' grades (from high to low).
2. Introduce a new type of crossover: a fixed-point crossover on grade-boundaries.
3. Split the population into several sub-populations based on grades
4. Introduce new (sub-)fitness functions based on a pseudo measure of under-covering for each grade.
5. Produce some of the children by applying the crossover operator to individuals from complementary sub-populations.

A sorted string can be regarded as three sub-strings - one for each grade, and the new crossover is restricted to crossover on the boundary points. The idea behind this new 'grade-based' crossover is to force the genetic algorithm to keep good sub-solutions or building blocks, based on the grades, together. To ensure a continued variety of different solutions only some of the crossovers are performed in this new way. The rest are done uniformly. However, as our objective is to avoid the problems of epistasis caused by nurses of other grades, the selection of parents cannot depend on the fitness of the whole string. Thus, additional sub-fitnesses according to the building blocks have to be introduced. This is achieved by splitting the population into a number of sub-populations and partitioning the covering constraints into 3 independent groups in order to define a pseudo measure of under-covering.

To do this we define new constants $r_{is}$ = 1 if nurse $i$ is of grade $s$, = 0 otherwise, and $S_{ks}$ as the demand on day/night $k$ for grade $s$ excluding that required at higher grades.

We then rewrite the constraints $\sum_{j=1}^{m}\sum_{i=1}^{n} q_{is} a_{jk} x_{ij} \geq R_{ks} \quad \forall k,s \quad (3)$

as:
$$\sum_{j=1}^{m}\sum_{i=1}^{n} r_{i1} a_{jk} x_{ij} \geq S_{k1} \quad \forall k \quad (3.1)$$

$$\sum_{j=1}^{m}\sum_{i=1}^{n} r_{i2} a_{jk} x_{ij} \geq S_{k2} \quad \forall k \quad (3.2)$$

$$\sum_{j=1}^{m}\sum_{i=1}^{n} r_{i3} a_{jk} x_{ij} \geq S_{k3} \quad \forall k \quad (3.3)$$

With $S_{ks} = \begin{cases} R_{ks} & s = 1 \\ R_{ks} - R_{k(s-1)} & s = 2,3 \end{cases}$

Note that (3.1), (3.2) and (3.3) will only match (3) if the covering constraints are tight at each grade. Otherwise, (3) allows higher graded nurses to cover requirements at lower grades. For example if there are more than enough grade 1 and grade 2 nurses and the overall cover is tight, constraints (3.1) and (3.2) will be slack, but constraints (3.3) will never be satisfied. Thus, these new constraints are not considered binding but are merely included to guide the search within the sub-populations. As violations are penalised in the fitness function, solutions that are close to providing the required cover at a particular grade will be fitter than those that are not.

These new constraints are used to guide the search in 7 sub-populations defined as follows:
Sub-populations 1, 2 and 3 have fitnesses based on cover and requests for grade 1, 2 and 3 respectively.
Sub-populations 4, 5, 6 and 7 have fitnesses based on cover and requests for grade 1+2, 1+3, 2+3 and 1+2+3 respectively.
For example the fitness function for population 1 would be:

$$\sum_{i \in \text{grade}1}\sum_{j=1}^{m} p_{ij} x_{ij} + g_{\text{demand}}(1) \sum_{k=1}^{14} \max\left[ S_{k1} - \left(\sum_{i \in \text{grade}1}^{n}\sum_{j=1}^{m} r_{i1} a_{jk} x_{ij}\right); 0 \right]$$

and that for population 4 would be:

$$\sum_{i \in \text{grade}1,2}\sum_{j=1}^{m} p_{ij} x_{ij} + g_{\text{demand}}(4) \sum_{k=1}^{14} \left\{ \begin{array}{l} \max\left[ S_{k1} - \left(\sum_{i \in \text{grade}1}^{n}\sum_{j=1}^{m} r_{i1} a_{jk} x_{ij}\right); 0 \right] \\ + \max\left[ S_{k2} - \left(\sum_{i \in \text{grade}2}^{n}\sum_{j=1}^{m} r_{i2} a_{jk} x_{ij}\right) -; 0 \right] \end{array} \right\}$$

As we are using dynamic weights, it is necessary to decide whether $g_{\text{demand}}$ should be based on the best individual over-all or should be calculated separately for each sub-population. Initial experiments confirmed our view that the latter gives better results and this was therefore adopted. Note that this results in different values for $g_{\text{demand}}(B)$ in each population $B$ at any point in time. Note also that population 7 does not represent the original optimisation problem as it is concerned only with the total cover and not with the cover at grades 1 and 2. It is therefore necessary to maintain a 'main' population - sub-population 8 whose fitness is the original fitness given in section 4.

In order to solve the original problem we also need to strike a balance between producing highly fit individuals with respect to sub-strings in the sub-populations and allowing individuals from complementary populations to combine co-operatively. This is achieved by the following rules:
1. Sub-populations 1-3 evolve by uniform crossover within themselves for maximum diversity.
2. Sub-populations 4-7 evolve by 50% uniform and 50% grade-based crossover. In the case of grade-based crossover the parents are picked from sub-populations 1-3 as appropriate and combined accordingly e.g. grade-based children in population 4 would have parents from 1 and 2.
3. Main population 8 evolves by 50% uniform and 50% grade-based crossover. In the case of grade-based crossover the parents are picked from sub-populations 1-7 and combined by crossover at the appropriate grade boundaries.

Experimentation led to the adoption of a population size of 100 for sub-populations 1-7, with the remaining 300 individuals belonging to population 8.

In spite of the exchange of information between species due to the grade-based crossover, some additional migration was found to be advantageous. Experiments with different forms of migration showed little difference in performance between the options and we opted for a simple random migration, in which an individual moves from one population to another, every few generations.

The results of this variant are illustrated in Figure 1b. Comparing this with Figure 1a, a vast improvement is apparent. The co-operating sub-populations lead to a more effective search by combining low cost feasible building blocks. All the instances but one are solved to feasibility at least once, and many solutions are now of high quality. The mean solution time is 17.33 seconds, implying that several random starts would be possible within a few minutes of total run time. However, there remain a few instances for which solution quality is still relatively poor. These are addressed in the following section.

### 7. Incentives and Local Search/Repair

A look at the failed attempts (i.e. those runs that failed to find either good quality or feasible solutions) showed that the search was converging towards solutions that satisfied most, but not all, of the covering constraints. These could be partitioned into two distinct types according to the way in which the over-covering and under-covering was distributed over the group of day shifts and the group of night shifts. We will denote these balanced and unbalanced. A balanced solution is essentially one that can be repaired by making small changes to the patterns worked by individual nurses, without moving any nurses from days to nights, or vice versa. An unbalanced solution is one where the 'wrong' nurses are on days and nights, but it is unlikely that the crossover operator will be able to produce fitter children. The two classes are defined formally as follows.

In a balanced solution either the group of day shifts or the group of night shifts are all covered perfectly without over-covering or under-covering on any shift. The imperfectly covered group has some shifts over-covered and some under-covered, such that the total over-covering is sufficient to compensate for all the under-cover. The first two rows of Table 1 illustrate typical balanced solutions. In an unbalanced solution, one of the groups has at least one under-covered shift, and at most one has over-covered shifts. If the same group has under and over-cover, then the over-cover should be insufficient to compensate for the under-cover. Examples of unbalanced solutions are given in rows 3 to 5 of Table 1. Note that situations such as that in row 4 are possible as most, but not all, nurses work less shifts when on days than on nights. Any other solutions are classified as undecided.

**Table 1**
**Balanced, unbalanced and undecided solutions**

The columns represent the 7 day and night shifts.
The number is each cell is the amount of over-covering

| Type | day | | | | | | | Night | | | | | | |
|---|---|---|---|---|---|---|---|---|---|---|---|---|---|---|
| | 1 | 2 | 3 | 4 | 5 | 6 | 7 | 1 | 2 | 3 | 4 | 5 | 6 | 7 |
| Balanced | -2 | 0 | 1 | 0 | 1 | 1 | 0 | 0 | 0 | 0 | 0 | 0 | 0 | 0 |
| Balanced | 0 | 0 | 0 | 0 | 0 | 0 | 0 | 0 | -1 | 0 | 0 | 0 | 1 | 0 |
| Unbalanced | 0 | 0 | -1 | 0 | 0 | 0 | 0 | 0 | 1 | 0 | 0 | 0 | 0 | 0 |
| Unbalanced | 0 | 0 | -1 | 0 | 0 | 0 | 0 | 0 | 0 | 0 | 0 | 0 | 0 | 0 |
| Unbalanced | 0 | 0 | -2 | 0 | -1 | 0 | 0 | 0 | 0 | 0 | 0 | 0 | 0 | 0 |
| Undecided | 0 | 0 | -1 | -1 | 0 | 2 | 0 | 0 | 0 | 1 | -1 | 0 | 0 | 0 |
| Undecided | 0 | -1 | -1 | 1 | 0 | 0 | -2 | 0 | 0 | 2 | 0 | 2 | 0 | -1 |

Although fit balanced solutions can usually be repaired by changing the patterns of a few nurses (e.g. in row one of the table it can be corrected by moving one nurse currently working day 3 to day 1 and one from day 5 to day 1), such small changes are difficult for the GA due to the disruptiveness of the crossover operator. Nevertheless runs that converged to balanced solutions usually terminated successfully if allowed to run for long enough, due to the use of the mutation operator. As nurses are usually contracted to work either days or nights in a given week such a mutation will not swap a day shift for a night shift. For example row 3 of the table cannot be corrected by moving a nurse from night 2 to day 3 as her other shifts that week will be nights. Thus, once the search has converged to solutions where the 'wrong' nurses are on days and nights it is difficult for the GA operators to sort the situation out.

The solution to this dilemma is to look ahead and avoid unbalanced solutions by penalising them more severely and rewarding solutions with a 'future' potential, i.e. balanced solutions. Therefore, fitness scores of balanced and unbalanced solutions are now adjusted by adding a bonus to balanced solutions and a disincentive or negative bonus to unbalanced solutions. This results in balanced, but less fit, solutions ranking higher than unbalanced fitter solutions, which is the desired effect. Both the bonus and penalty are based on the weights used for the constraint violation penalties, multiplied by a constant factor. Thus they change dynamically at the same rate. Experiments using a variety of weights were carried out and a weight of 3 x constraint violation penalty was found to give consistently good results.

Logically, the result of the above change will result in more runs converging towards balanced solutions. As discussed above many of these eventually become feasible as a result of random mutation. It makes sense to exploit this situation by introducing a more intelligent mutation operator that will attempt to repair these solutions directly. This will not only reduce the number of generations before such solutions become feasible, but will also counter the disruptiveness inherent in the crossover operator and ensure that such solutions reach their full potential before being destroyed by crossover. The repair operator is a simple hill-climber and works as follows. A balanced solution is taken and subjected to an improvement heuristic that cycles through each nurses' shift patterns, accepting a new pattern if it improves fitness. This results in the ability to make only slight changes in the worked days/nights of a single nurse. However, this routine is only applied to the top five solutions. This avoids potential problems of premature convergence resulting from using too much aggressive mutation, and saves wasting computation time repairing/improving large numbers of balanced solutions in later generations. The hill-climber is also applied to feasible solutions in which case it aims to improve the preferences.

| Table 2<br>Performance of incentives and repair in terms of feasibility and solution quality ||||||
|---|---|---|---|---|---|
| All figures are based on 20 runs over 52 data sets. ||||||
| type of incentive | None | incentive only | incentive and repair | disincentive only | both and repair |
| % feasible | 75% | 78% | 83% | 87% | 89% |
| Mean total $p_{ij}$ | 17.6 | 14.3 | 12.1 | 13.5 | 10.9 |

Table 2 illustrates the effects of the different options. Note how the incentive alone has only limited impact and needs the hill-climber to improve solution quality. The disincentive alone improves feasibility but not so much quality. The best results are obtained when all three features are present. Results on an individual basis are given in Figure 1c. All problems now have at least one feasible solution and most fall acceptably close to the optimum. Solution times have also reduced to 14.9 seconds per run.

It is difficult to make a direct comparison between these results and those of the tabu search, as the implementation details are very different. When run with the 'optimum' choice of parameters the latter takes between 3 and 4 times as long as the genetic algorithm, but solution quality is better and more consistent. However, when the number of iterations is reduced to match the times quoted above, quality and consistency are degraded to similar levels to those of the full GA (except in situations where the GA finds few feasible solutions when the tabu search is better). As the tabu search is specifically geared to seek out feasible solutions it rarely fails in this respect.

When the tabu search is run without the special moves, solution quality falls below that of the GA, with a mean total $p_{ij}$ value of 13.2. It is also worth noting that those problems where the GA frequently fails to find feasible solutions are those in which the penalty cost of the optimal solution is far higher than the best solution with just one covering constraint violation. This suggests that the severity parameter $\alpha = 8$ may not be high enough for such problems. A policy in which $\alpha$ is increased if the first few runs fail to find any feasible solutions may help to overcome this.

In order to gauge the flexibility of the algorithm with respect to the $p_{ij}$ values, both the GA and tabu search were run on 52 further instances with randomly generated $p_{ij}$. Both methods were less consistent than on the original data, and neither completely dominated the other. However, as expected the advantage of the special chain moves in the tabu search was much reduced, and on average the GA was able to produce better results in less computation time. We therefore conclude that, although the full tabu search is the better of the two for instances with solution landscapes similar to those of the original 52 problems, the GA scores for speed, simplicity and robustness.

## 8. Conclusions

This paper has shown how the underlying structure of a nurse scheduling problem can be exploited in a genetic algorithm approach to the problem. This knowledge was used in two ways. The first uses the grade-based structure to define a hierarchy of sub-populations in order to build good partial solutions. These were combined strategically using crossover to produce children for populations further up the hierarchy. Although this was a vast improvement over a canonical GA there were still some problem instances where the algorithm had difficulty in finding feasible solutions. These failures appeared to be due to the well-known problem that GAs 'lack the killer instinct' [6] in that they converge towards good solutions, but are too disruptive to make the minor changes necessary to improve these. This problem is frequently overcome by adding a hill-climber to make small improvements to some or all of the population. Here we improved the success rate of this strategy by recognising solution attributes that were likely to lead to successful hill-climbing, and those that certainly would not, rewarding the former class and penalising the latter.

The result is a fast robust implementation, in which all the add-ons are relatively simple and remain very much within the spirit of the GA. It has proved amenable to additional objectives and constraints as required by the hospital from time to time and empirical evidence from tests with larger numbers of nurses, different distributions of penalty weights and different numbers of grade bands confirm its flexibility.

As the additional information is, by definition, problem specific these enhancements cannot be copied directly into a GA implementation for a different problem. Nevertheless the underlying concepts may prove useful in other scenarios. Many scheduling problems have objectives or constraints that involve subsets of elements. The idea of co-evolving, co-operative sub-populations forming large building blocks that meet specific aspects of the problem could prove useful in these circumstances. As most scheduling problems have natural local search neighbourhoods a hill-climber to repair or improve good solutions can usually be applied. The idea of incentives and disincentives depends on the presence and easy detection of good and bad attributes, but our success with this modification suggests that it is certainly worth consideration.

However, we do not claim that this is the necessarily the best GA approach to the problem. Although simpler than the problem specific add-ons in the tabu search approach, the development effort is considerable. In an early paper Davis [5] suggests an indirect coding combined with a heuristic decoder as an effective means of overcoming the problems of epistatic domains, and since then this type of GA has been used successfully on a variety of problems. This approach has the advantage that all the power of the problem specific information is contained within the heuristic, while a basic GA works to optimise its parameters, for example the order in which the nurses are to be processed. We are currently developing such a GA for the nurse rostering problem with the objective of comparing the two approaches both in terms of solution time and quality, and development effort.